\documentclass[letterpaper]{article} 
\usepackage[preprint]{aaai2027}  
\usepackage[hyphens]{url}  
\usepackage{graphicx} 
\usepackage{natbib}  
\usepackage{caption}
\usepackage{makecell}
\usepackage{amsmath}
\usepackage{amssymb}
\usepackage{booktabs}
\usepackage{multirow}
\usepackage{pifont}
\usepackage{bm} %
\title{FreqNav: Stage-Wise Frequency Routing for \\ Object-Oriented Aerial Vision-Language Navigation}
\author{
Yin Tang$^1$,
Jiawei Ma$^2$,
Jiahao Li$^3$,
Hao Zhang$^4$,
Zhemin Sun$^3$,
Jianqiao Sun$^4$,
Deyu Zhang$^3$
}

\affiliations{
$^1$ Big Data Institute, Central South University\\
$^2$ Department of Computer Science, City University of Hong Kong\\
$^3$ School of Computer Science and Engineering, Central South University\\
$^4$ School of Electronic Engineering, Xidian University\\
}
\begin{document}
\maketitle
\begin{abstract}
Object-oriented aerial vision-and-language navigation (VLN) requires searching for a described target and landing on it precisely, under long-horizon and closed-loop control. 
Guided by a target‑descriptive instruction during navigation, perceptual priorities dynamically evolve: early‑stage exploration prioritizes low‑frequency spatial layout, and then shifts to high‑frequency target details. 
Existing VLN methods model the varying perceptual requirements across navigation stages with identical visual tokens, leading to interference from irrelevant objects and background clutter.
To this end, we therefore formulate long-horizon aerial navigation as a frequency-preference shift from spatial structure to local detail and propose \textbf{FreqNav}, a lightweight frequency-routing adaptive perception framework. Under a fixed computational budget, FreqNav dynamically reallocates visual tokens across frequency components according to the current navigation stage. A Frequency Token Router selects stage-relevant visual representations from dual-view observations, while a Phase-dependent Grounding Module anchors visual evidence through explicit supervision. A Diffusion Transformer then predicts smooth trajectories for continuous control. Experiments show that FreqNav outperforms strong baselines while achieving approximately 3× faster inference. Real-world deployment further demonstrates its effectiveness, efficiency, and practical potential for long-horizon aerial autonomy.
\end{abstract}

\begin{links}
    \link{Demo Link}{https://youtu.be/zS_NjWQ2X0U}.
\end{links}

\section{Introduction}
\label{sec:introduction}

Object-oriented aerial vision-language navigation (VLN) requires unmanned aerial vehicles (UAVs) to autonomously perform target search and precise landing under the guidance of visual observations and natural language instructions~\cite{fan2023avdn, gao2025geonav, xiao2025uav}. This capability is of great practical significance for urban search and rescue~\cite{maresca2025react}, logistics delivery~\cite{zhang2025logisticsvln}, and automated inspection~\cite{yang2025det}.
Unlike indoor navigation, outdoor aerial navigation in unstructured environments calls for a dual capability: active semantic exploration and spatial-geometric perception when the target is not visible, and fine-grained control for accurate landing as the UAV approaches the target.
In large-scale open scenarios, such a task poses unique challenges to the agent’s ability to flexibly switch between global exploration and local target approach across navigation stages. 

\begin{figure}[t]
    \centering
    \includegraphics[width=1\linewidth]{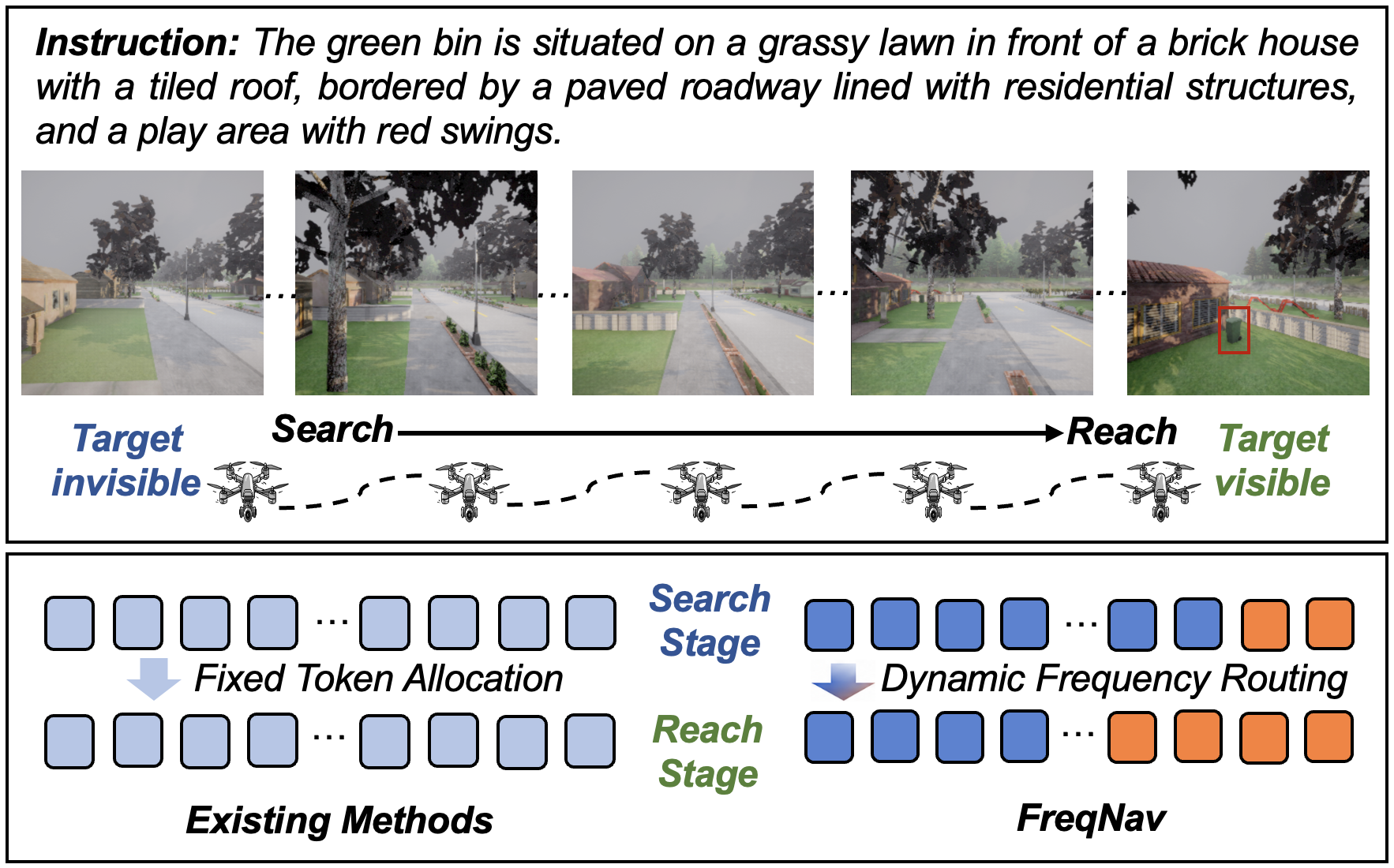}
    \caption{\textbf{Motivation of FreqNav}. Aerial VLN is split into the search stage (target invisible) and reach stage (target visible). While prior works use static uniform visual token allocation, our FreqNav enables dynamic frequency routing. It realizes an object-conditioned frequency preference shift to adaptively trade between low-frequency global scene representations and high-frequency target details for different navigation phases.}
\label{fig:freq}
\end{figure}

Recent aerial VLN models primarily rely on recurrent policies with cross-modal attention~\cite{liu2023aerialvln}, explicit long-horizon navigation architectures~\cite{wang2025traveluav, gao2025geonav, ding2026history}, or end-to-end vision-language-action (VLA) models~\cite{liu2023aerialvln}. 
While these methods have significantly advanced Aerial VLN, they predominantly employ a static visual perception paradigm that uniformly allocates visual tokens across all navigation stages. In contrast, object-oriented navigation exhibits stage-disparate perceptual requirements: the initial exploration prioritizes global scene layouts and spatial geometry, whereas the final reach relies on localized target appearance and fine-grained semantic alignment. 
As established in prior studies~\cite{xu2020frequency,qi2026fuse,wang2026fouriercompressor}, global structural scene layouts are primarily encoded in low-frequency signal components, while fine-grained object details predominantly reside in high-frequency components. We argue that stage-wise adaptive token selection can yield information-denser visual representations by prioritizing frequency components most relevant to the current phase. As shown in Figure \ref{fig:freq}, this frequency preference shift across navigation stages motivates us to conceptualize long-horizon aerial navigation as a dynamic frequency reallocation process, necessitating a perception framework that adaptively routes frequency-aware visual representations throughout the navigation trajectory.

To this end, we propose FreqNav, a frequency-routing adaptive perception framework for long-horizon aerial visual language navigation. Specifically, we firstly introduce the Frequency Token Router, which adaptively selects stage-relevant visual tokens via frequency representations from dual-view observation. Such adaptive reallocation not only enhances representation fidelity for more accurate decision-making but also enables compression of redundant tokens to substantially improve inference efficiency. On the selected tokens, we then design a Phase-dependent Grounding module to anchor visual cues for explicit supervision. Finally, a Diffusion Transformer (DiT)-based action generation is conditioned on the grounded query states to predict smooth trajectories for robust closed-loop flight control.

Extensive experiments comprehensively validate the superiority of FreqNav in both accuracy and inference efficiency. On the TravelUAV benchmark, FreqNav achieves a 51.55\% Success Rate (SR) and 41.67\% SPL in Seen Scenarios, 39.04\% SR and 30.50\% SPL on Unseen Maps, and 57.23\% SR and 47.00\% SPL on Unseen Objects, consistently outperforming the strongest end-to-end VLA baseline. Simultaneously, by effectively compressing the standard 128-token visual representation down to 50 routed tokens, our model achieves a 3× inference speed-up. Furthermore, an air-ground collaborative deployment on the Feisi J310 UAV platform demonstrates that the system operates stably with a server-side inference latency of 0.17 s/step, verifying its practicality and reliability for real-world long-horizon aerial autonomy. Our contribution is threefold.
\begin{itemize}
    \item We formulate object‑oriented aerial VLN as a transition from target‑search and target‑reach phases, and map this to a frequency‑preference shift from global spatial layout reasoning to local target‑detail grounding.
    
    \item We propose FreqNav, a lightweight frequency-routing adaptive perception framework. FreqNav uses a Frequency Token Router to construct compact frequency-aware visual representations from dual-view observations, a Phase-dependent Grounding module to provide explicit spatial supervision, and a  Diffusion Transformer to predict smooth continuous trajectories.  
    \item FreqNav achieves state-of-the-art navigation accuracy and efficiency on the TravelUAV benchmark. Furthermore, real-world deployment via an air-ground collaboration validates FreqNav as a feasible solution for embodied intelligent UAVs navigating open scenarios.  
\end{itemize}

\section{Related Work}
\label{sec:related_work}

\subsection{Aerial Vision-Language Navigation}

Vision-language navigation began with instruction following in indoor and street-view environments, where agents align language with visual observations and choose actions over discrete or continuous spaces~\cite{anderson2018vision,chen2019touchdown,krantz2020continuous}.
Aerial VLN extends this problem to large-scale 3D motion, longer trajectories, sparse target visibility, and continuous control.
AerialVLN and Aerial Vision-and-Dialog Navigation established early UAV instruction-following settings~\cite{liu2023aerialvln,fan2023avdn}, while CityNav introduced language-goal aerial navigation with geographic context~\cite{lee2024citynav}.
TravelUAV and its UAV-Need-Help task provide realistic AirSim trajectories and official seen, unseen-object, and unseen-map splits~\cite{wang2025traveluav}.
GeoNav and CityNavAgent add explicit geospatial reasoning, hierarchical semantic planning, and global memory~\cite{gao2025geonav,zhang2025citynavagent}.

Recent methods increasingly target autonomy rather than dense oracle assistance. For example, AerialVLA uses a dual-view VLA policy with fuzzy onboard directional prompts and an intrinsic landing decision~\cite{xu2026aerialvla}.
HETT combines coarse target prediction, fine action refinement, and a historical grid map~\cite{ding2026history}.
CityAVOS and PRPSearcher formulate urban object search as an exploration and exploitation process over multiple 3D maps~\cite{ji2026cityavos}. Although effective, existing methods formulate aerial navigation as a static perception process, using a fixed visual token allocation for both target search and target reach. In contrast, FreqNav dynamically reallocates stage-aware visual tokens and further employ a diffusion-style policy~\cite{chi2025diffusion,lipman2023flowmatching} to generate smooth trajectories.

\subsection{Frequency-Domain Networks}
Frequency-domain learning uses Fourier or cosine transforms to encode visual features into hierarchical frequency components, enabling the modeling of global and local visual information. To obtain hierarchical feature representations, recent methods have explicitly decompose multimodal features into different frequency components, associating low-frequency signals with global scene structure and higher frequencies with fine-grained local details~\cite{xu2020frequency,qi2026fuse}. Meanwhile, visual token optimization methods improve efficiency through token merging, pruning, or query-based compression to remove redundant visual information~\cite{bolya2023tome,li2023blip2,wang2026fouriercompressor,mnih2014attention}. However, these approaches assume a static frequency representation or token allocation, overlooking the evolving perceptual demands of embodied navigation. In contrast, FreqNav is the first to formulate aerial navigation as a frequency-preference transition process, dynamically routing stage-adaptive frequency representations to produce information-dense visual features.
\section{Method}
\label{sec:method}

\begin{figure*}[t]
    \centering
    \includegraphics[width=1\textwidth]{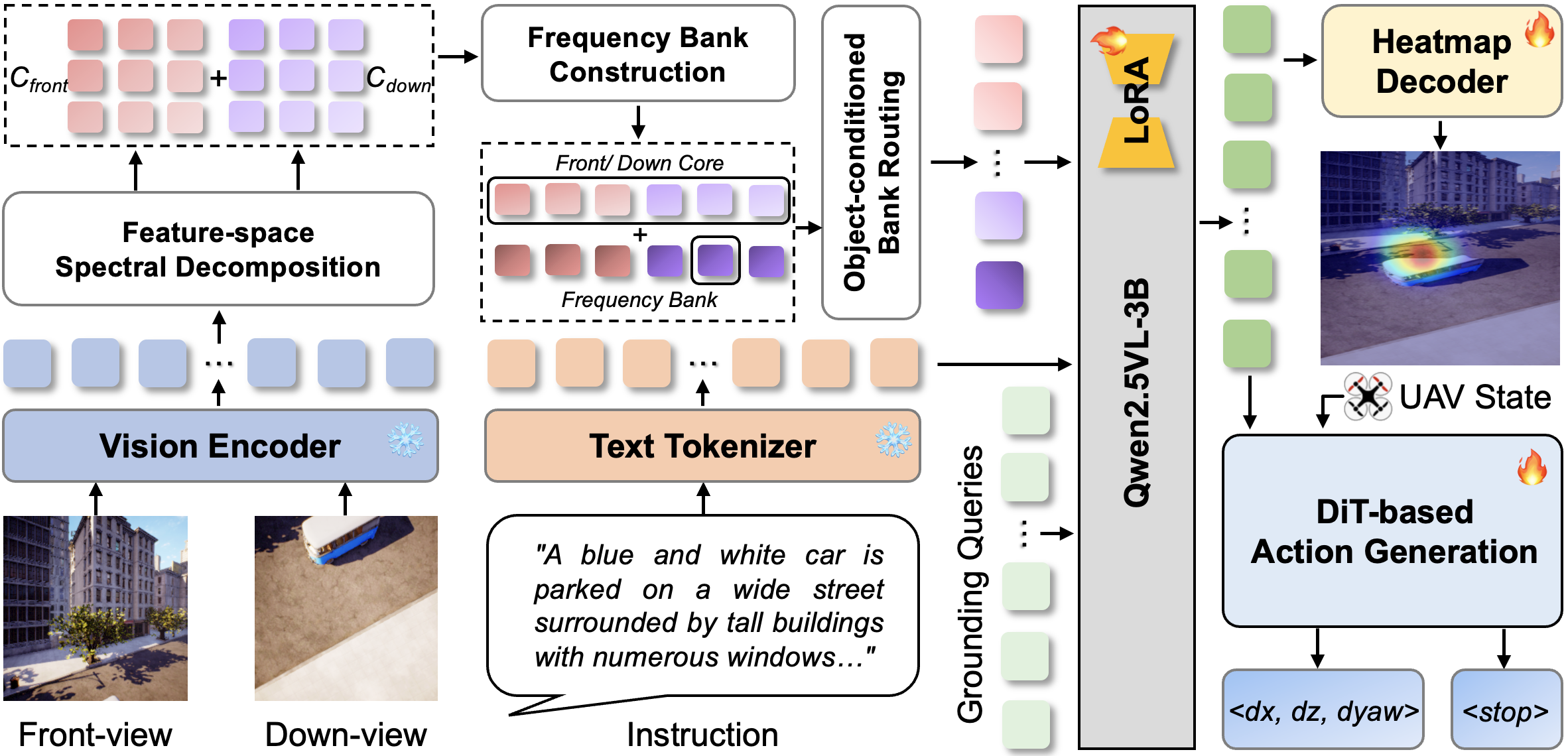}
    \caption{\textbf{Overview of FreqNav.} At each time step $t$, a frozen vision encoder extracts front- and down-view features, which are fed into \textit{Frequency Token Router} to get routed visual tokens. Then, the \textit{Phase-dependent Grounding} jointly processes the routed visual tokens, instruction tokens, and learnable grounding queries through VLM to obtain grounded query states, with a training-only heatmap decoder providing explicit spatial supervision. Conditioned on the grounded query states and UAV state, we employ \textit{DiT-based action generation} to predict the continuous waypoint $\langle dx,dz,d\mathrm{yaw}\rangle$ together with a stop probability for closed-loop aerial navigation.}
    \label{fig:Framework}
\end{figure*}

\paragraph{Problem Formulation.}
We formulate object-oriented aerial VLN as a partially observable Markov decision process. At time $t$, the UAV agent receives: 
\begin{equation}
o_t=\left(I_t^{f},I_t^{d},g,\boldsymbol{\eta}_t,\mathbf{p}_t\right),
\end{equation}
where $I_t^{f}$ and $I_t^{d}$ are front- and down-view RGB images, $g$ is the natural-language target description, $\boldsymbol{\eta}_t$ is a step-wise direction prior that indicates the relative bearing of the target~\cite{xu2026aerialvla}, and $\mathbf{p}_t\in\mathbb{R}^{6}$ is the current UAV state.
Given the observation $o_t$, the UAV agent predicts a $K$-step continuous waypoint chunk $\mathbf{A}_t=[\mathbf{a}_{t,1},\ldots,\mathbf{a}_{t,K}]$, where $\mathbf{a}_{t,k}=[a_{t,k}^{\mathrm{fwd}},a_{t,k}^{\mathrm{down}},a_{t,k}^{\mathrm{yaw}}]\in\mathbb{R}^{3}$, together with a stop probability $\hat{s}_t$.
Let $\mathbf{x}^{\star}$ denote the ground-truth target position and $\mathbf{x}_T$ the UAV's landing position, with stop-decision threshold $\delta_{\mathrm{stop}}$, an episode is successful when the UAV agent terminates within the maximum horizon $T_{\max}$ and stops within the success radius $d_{\mathrm{succ}}=20\,\mathrm{m}$.

\paragraph{Model Overview.}
FreqNav constructs the navigation policy by modeling aerial VLN as a frequency-preference transition from target search to target reach. At each time step, FreqNav first use the Qwen2.5-VL vision encoder~\cite{bai2025qwen25vl} to encode the vertically stacked front- and down-view RGB images into visual tokens. The \ding{172} \textit{Frequency Token Router $\mathcal{R}$} then allocates a fixed visual token budget across frequency components, preserving low-frequency scene cores from both views while dynamically selecting a object-conditioned evidence bank as the observation evolves from search to reach. Given the routed tokens, we further design the \ding{173} \textit{Phase-dependent Grounding $\mathcal{G}$} to anchor the visual cues for explicit supervision. Finally, a \ding{174} \textit{DiT-based Action Generation $\mathcal{D}$} conditions on the grounded tokens to generate the continuous waypoints and stop probability, enabling closed-loop replanning toward smoother and more stable long-horizon flight. Next, we will detail the each module in the following.

\subsection{Frequency Token Router}
\label{sec:router}
To dynamically select target-relevant visual evidence during different stages, we introduce the Frequency Token Router. It first performs feature-space spectral decomposition and extracts a fixed low-frequency front/down core to preserve global semantics. It then constructs adaptive evidence banks from spectral reconstructions, providing complementary local cues. Finally, a object-conditioned discrete router selects one evidence bank and combines it with the fixed core for grounding.

\paragraph{Feature-space Spectral Decomposition.}
Given front and down views $v\in\{f,d\}$, we apply the frequency transform to visual feature grids rather than to raw RGB pixels. For every feature channel, an orthonormal two-dimensional DCT is applied along the two spatial axes:
\begin{equation}
\mathbf{C}_t^{v}=\mathbf{D}\mathbf{F}_t^{v}\mathbf{D}^{\top},
\end{equation}
where $\mathbf{F}_t^{v}\in\mathbb{R}^{B\times8\times8\times H}$ is the view-specific visual grid and the transform leaves the $H$ channels independent.
This transform exposes spatial-frequency coordinates over the VLM token lattice while preserving the channel dimension, so later selection can operate in frequency space and still return spatial feature tokens to the language backbone.

\paragraph{Frequency Bank Construction.}
After decomposition, we extract a fixed low-frequency core and adaptive evidence bank for frequency bank construction. For the  fixed core, we apply a low-pass crop $\Pi_{\mathrm{lp}}$ to each view and reconstructing the result back to the spatial feature domain:
\begin{equation}
\bar{\mathbf{F}}^{v}_{t,\mathrm{lp}}=
\operatorname{Flatten}
\!\left[
\operatorname{IDCT}\!\left(\Pi_{\mathrm{lp}}(\mathbf{C}_t^{v})\right)
\right]
\in\mathbb{R}^{B\times N_{\mathrm{core}}^{v}\times H},
\end{equation}
where $N_{\mathrm{core}}^{v}$ is the fixed token budget assigned to view $v$. Based on this, we obtain the reconstructed front/down cores  $\mathbf{T}^{f}_{\mathrm{core}}$ and $\mathbf{T}^{d}_{\mathrm{core}}$ that preserve global semantics as grounding cues.

For the adaptive banks, we construct them from view- and band-specific spectral reconstructions. Let $r(i,j)=\max(i,j)$ denote the frequency radius on the $8\times8$ grid, and define three overlapping spectral masks:
\begin{equation}
\mathcal{M}_{\mathrm{low}}:r\leq2,\qquad
\mathcal{M}_{\mathrm{mid}}:2\leq r\leq4,\qquad
\mathcal{M}_{\mathrm{high}}:r\geq4.
\end{equation}
For each view-band pair $(v,b)$, where $v\in\{f,d\}$ and $b\in\{\mathrm{low},\mathrm{mid},\mathrm{high}\}$, we mask the coefficients, reconstruct them back to the spatial feature domain, and resample the result into an ordered adaptive bank:
\begin{equation}
\mathbf{B}_{v,b}=
\operatorname{Flatten}
\left[
\operatorname{IDCT}
\left(\mathcal{M}_{b}\odot\mathbf{C}_t^{v}\right)
\right]
\in\mathbb{R}^{B\times N_{\mathrm{bank}}\times H}.
\end{equation}
This yields six dynamic frequency banks from the Cartesian product of two views and three spectral bands. Low-band banks supplement coarse layout, mid-band banks emphasize object-scale structure, and high-band banks retain localized boundary and texture-like evidence. We also set the low/mid masks overlap at $r=2$ and the mid/high masks overlap at $r=4$ to reduce sensitivity of hard band boundaries.

\paragraph{Object-conditioned Bank Routing.}
To dynamically isolate phase-specific visual cues within a fixed token budget, we introduce a discrete routing mechanism. Given the text embedding of the target description $\mathbf{e}_g$, we directly project it into unnormalized logits $\boldsymbol{l}_t \in \mathbb{R}^6$, representing the routing preferences over the six candidate frequency banks $\{\mathbf{B}_i\}_{i=1}^6$. 

To enable end-to-end differentiable training for this discrete selection, we employ the straight-through Gumbel-Softmax estimator~\cite{jang2017categorical}. The soft selection probability for the $i$-th bank is computed as:
\begin{equation}
z_{t,i} = \frac{\exp((l_{t,i} + \epsilon_i) / \tau)}{\sum_{j=1}^{6} \exp((l_{t,j} + \epsilon_j) / \tau)},
\end{equation}
where $\epsilon_i \sim \mathrm{Gumbel}(0,1)$ is i.i.d. noise and $\tau$ is the temperature. During the forward pass, we enforce a strict discrete choice via $\hat{\mathbf{z}}_t = \mathrm{one\_hot}(\arg\max_i z_{t,i})$ to select the adaptive bank $\mathbf{B}_{t,\mathrm{adapt}} = \sum_{i=1}^{6} \hat{z}_{t,i} \mathbf{B}_i$, while retaining the continuous gradients of $\mathbf{z}_t$ during the backward pass. 

Finally, this object-conditioned adaptive bank is concatenated with the fixed low-frequency cores to construct the final routed visual tokens:
\begin{equation}
\mathbf{T}_t^{\mathcal{R}} = \left[\mathbf{T}^{f}_{\mathrm{core}}, \mathbf{T}^{d}_{\mathrm{core}}, \mathbf{B}_{t,\mathrm{adapt}}\right].
\end{equation}
Through this routing mechanism, FreqNav dynamically shifts its visual evidence preference from global semantic reasoning to local target detail capture, yielding a refined visual representation for the subsequent Phase-dependent Grounding $\mathcal{G}$.

\subsection{Phase-dependent Grounding}
\label{sec:grounding}

To explicitly ground the dynamically routed visual tokens into actionable navigational cues, we introduce a query-based spatial grounding mechanism. The routed visual tokens $\mathbf{T}_t^{\mathcal{R}}$ are first inserted into the prompt sequence of the Qwen VLM. Following these tokens, we append $N_q=128$ learnable grounding queries, which are arranged as a $16\times8$ grid and injected with two-dimensional positional encodings to preserve spatial topology. Through the deep self-attention layers of the VLM, these learnable queries interact extensively with the phase-aware visual features and the target description. The VLM outputs the corresponding final hidden states for these queries, denoted as:
\begin{equation}
\mathbf{Q}_t \in \mathbb{R}^{B \times 128 \times H}.
\end{equation}

To enforce explicit spatial grounding on $\mathbf{Q}_t$, we apply auxiliary prediction heads to output a 2D spatial heatmap. We consistently anchor the 2D heatmap to a short-horizon navigational target: the expert waypoint $h_g=3$ steps ahead, projected onto the stacked front/down views. By explicitly forcing the learnable queries to localize this future waypoint, the module effectively distills the phase-selected visual evidence into a concrete spatial representation, paving the way for the subsequent continuous action generation.

\subsection{DiT-based Action Generation}
\label{sec:dit}

To generate smooth and continuous waypoints from the spatially grounded representations, we adopt a Flow Matching framework parameterized by a Diffusion Transformer (DiT)~\cite{peebles2023dit}. 

\paragraph{Waypoint Generation.}
Formally, given $\mathbf{Q}_t$ from the grounding module and the current UAV state $\mathbf{p}_t$, the DiT directly models the continuous waypoint generation as a flow matching process. Let $\mathbf{A}^{\star}$ denote the expert $K$-step continuous waypoint chunk, we sample standard Gaussian noise $\boldsymbol{\epsilon} \sim \mathcal{N}(\mathbf{0},\mathbf{I})$ and a uniform flow time $\tau \sim \mathcal{U}(\varepsilon, 1-\varepsilon)$. The linear probability path $\mathbf{X}_{\tau}$ and its corresponding target velocity field $\mathbf{V}^{\star}$ are defined as:
\begin{equation}
\mathbf{X}_{\tau} = (1-\tau)\boldsymbol{\epsilon} + \tau\mathbf{A}^{\star}, \qquad \mathbf{V}^{\star} = \mathbf{A}^{\star} - \boldsymbol{\epsilon}.
\end{equation}
Through cross-attention mechanisms, the network utilizes the joint conditional context $[\mathbf{Q}_t, \mathbf{p}_t, \mathbf{h}_t]$ to predict the vector field $\mathbf{V}_{\theta}(\mathbf{X}_{\tau}, \tau \mid \mathbf{Q}_t, \mathbf{p}_t, \mathbf{h}_t)$ that matches $\mathbf{V}^{\star}$. Simultaneously, a parallel classification head is appended to the DiT to output the step-wise stop probability $\hat{s}_t$.

\paragraph{Training and Inference.}

To ensure stability and effectively decouple visual reasoning from continuous control, FreqNav is trained in a two-stage paradigm. In the first stage, we fine-tune the Qwen backbone via LoRA~\cite{hu2022lora} for high-level multi-modal perception. In the second stage, we freeze the VLM features and train the DiT controller for continuous waypoint generation.

During inference, the continuous waypoint generation starts from pure noise $\mathbf{X}_0 \sim \mathcal{N}(\mathbf{0},\mathbf{I})$ and use 16 uniform Euler steps to obtain $\mathbf{X}_1$, which is then clipped and denormalized into the physical simulator command range. To prevent the accumulation of open-loop execution errors, we empirically adopt a high-frequency replanning strategy, regenerating the action immediately after each executed step.

\begin{table*}[t]
\centering
\resizebox{\textwidth}{!}{
\begin{tabular}{l cccc cccc cccc}
\toprule
\multirow{2}{*}{Method} & \multicolumn{4}{c}{\textbf{Full}} & \multicolumn{4}{c}{\textbf{Easy}} & \multicolumn{4}{c}{\textbf{Hard}} \\
\cmidrule(lr){2-5} \cmidrule(lr){6-9} \cmidrule(lr){10-13}
& NE$\downarrow$ & SR$\uparrow$ & OSR$\uparrow$ & SPL$\uparrow$ 
& NE$\downarrow$ & SR$\uparrow$ & OSR$\uparrow$ & SPL$\uparrow$ 
& NE$\downarrow$ & SR$\uparrow$ & OSR$\uparrow$ & SPL$\uparrow$ \\
\midrule
Human & 14.15 & 94.51 & 94.51 & 77.84 & 11.68 & 95.44 & 95.44 & 76.19 & 17.16 & 93.37 & 93.37 & 79.85 \\
\midrule
Random Action & 222.20 & 0.14 & 0.21 & 0.07 & 142.07 & 0.26 & 0.39 & 0.13 & 320.12 & 0.00 & 0.00 & 0.00 \\
Fixed Action & 188.61 & 2.27 & 8.16 & 1.40 & 121.36 & 3.48 & 11.48 & 2.14 & 270.69 & 0.79 & 4.09 & 0.49 \\
CMA & 135.73 & 8.37 & 18.72 & 7.90 & 84.89 & 11.48 & 24.52 & 10.68 & 197.77 & 4.57 & 11.65 & 4.51 \\
TravelUAV-DA & 98.66 & 17.45 & 48.87 & 15.76 & 66.40 & 20.26 & 51.23 & 18.10 & 138.04 & 14.02 & 45.98 & 12.90 \\
NavFoM & 93.05 & 29.17 & 49.24 & 25.03 & 58.98 & 32.91 & 53.16 & 27.87 & 143.83 & 23.58 & 43.40 & 20.80 \\
LongFly & \textbf{60.02} & 36.39 & \textbf{65.87} & 31.07 & \textbf{38.10} & 38.52 & \textbf{71.90} & 31.24 & \textbf{85.20} & 33.94 & \underline{58.94} & 30.88 \\
AeroVLA & 65.88 & \underline{47.96} & 57.69 & \underline{38.54} & \underline{43.76} & \underline{49.30} & \underline{61.30} & \underline{37.14} & 93.16 & \underline{46.30} & 53.23 & \underline{40.26} \\
\midrule
\textbf{FreqNav (Ours)} & \underline{64.16} & \textbf{51.55} & \underline{60.79} & \textbf{41.67} & 45.64 & \textbf{49.81} & 60.54 & \textbf{38.04} & \underline{86.99} & \textbf{53.70} & \textbf{61.10} & \textbf{46.14} \\
\bottomrule
\end{tabular}
}
\caption{Comparison on the Test Seen Set. SR, OSR, and SPL are reported in percentage (\%). Bold and underline denote the best and second-best model results, respectively.}
\label{tab:seen_full_comparison}
\end{table*}
\section{Experiments}
\label{sec:experiments}
In this section, we evaluate our FreqNav on the TravelUAV benchmark~\cite{wang2025traveluav}, and detail the implementation protocols and main results.

\begin{table*}[t]
\centering
\resizebox{\textwidth}{!}{
\begin{tabular}{l cccc cccc cccc}
\toprule
\multirow{2}{*}{Method} & \multicolumn{4}{c}{\textbf{Full}} & \multicolumn{4}{c}{\textbf{Easy}} & \multicolumn{4}{c}{\textbf{Hard}} \\
\cmidrule(lr){2-5} \cmidrule(lr){6-9} \cmidrule(lr){10-13}
& NE$\downarrow$ & SR$\uparrow$ & OSR$\uparrow$ & SPL$\uparrow$ 
& NE$\downarrow$ & SR$\uparrow$ & OSR$\uparrow$ & SPL$\uparrow$ 
& NE$\downarrow$ & SR$\uparrow$ & OSR$\uparrow$ & SPL$\uparrow$ \\
\midrule
Random Action & 260.14 & 0.16 & 0.16 & 0.16 & 174.10 & 0.48 & 0.48 & 0.48 & 302.96 & 0.00 & 0.00 & 0.00 \\
Fixed Action & 212.84 & 3.66 & 9.54 & 2.16 & 151.66 & 6.70 & 13.88 & 3.72 & 243.29 & 2.14 & 7.38 & 1.38 \\
CMA & 155.79 & 9.06 & 16.06 & 8.68 & 102.92 & 14.83 & 22.49 & 13.90 & 182.09 & 6.19 & 12.86 & 6.08 \\
TravelUAV  & 118.11 & 22.42 & 46.90 & 20.51 & 86.12 & 24.40 & 49.28 & 22.03 & 134.03 & 21.43 & 45.71 & 19.75 \\
NavFoM  & 108.04 & 29.83 & 47.99 & 27.20 & 70.51 & 32.54 & 50.72 & 29.54 & 133.01 & 28.03 & 46.18 & 25.64 \\
LongFly & 66.74 & 43.87 & 64.56 & 38.39 & 54.84 & 38.01 & 56.84 & 31.36 & \textbf{57.07} & 50.25 & \textbf{74.16} & 45.27 \\
AeroVLA & \underline{61.45} & \underline{56.60} & \underline{64.86} & \underline{46.61} & \underline{45.72} & \textbf{56.94} & \underline{64.11} & \underline{43.76} & 69.27 & \underline{56.43} & 65.24 & \textbf{48.03} \\
\midrule
\textbf{FreqNav (Ours)} & \textbf{58.39} & \textbf{57.23} & \textbf{70.75} & \textbf{47.00} & \textbf{36.96} & \textbf{56.94} & \textbf{72.73} & \textbf{45.02} & \underline{69.06} & \textbf{57.38} & \underline{69.76} & \underline{47.99} \\
\bottomrule
\end{tabular}
}
\caption{Comparison on the Test Unseen Object Set. SR, OSR, and SPL are reported in percentage (\%). Bold and underline denote the best and second-best model results, respectively.}
\label{tab:unseen_obj_full_comparison}
\end{table*}

\begin{table*}[t]
\centering
\resizebox{\textwidth}{!}{
\begin{tabular}{l cccc cccc cccc}
\toprule
\multirow{2}{*}{Method} & \multicolumn{4}{c}{\textbf{Full}} & \multicolumn{4}{c}{\textbf{Easy}} & \multicolumn{4}{c}{\textbf{Hard}} \\
\cmidrule(lr){2-5} \cmidrule(lr){6-9} \cmidrule(lr){10-13}
& NE$\downarrow$ & SR$\uparrow$ & OSR$\uparrow$ & SPL$\uparrow$ 
& NE$\downarrow$ & SR$\uparrow$ & OSR$\uparrow$ & SPL$\uparrow$ 
& NE$\downarrow$ & SR$\uparrow$ & OSR$\uparrow$ & SPL$\uparrow$ \\
\midrule
Random Action & 202.98 & 0.00 & 0.00 & 0.00 & 158.46 & 0.00 & 0.00 & 0.00 & 265.88 & 0.00 & 0.00 & 0.00 \\
Fixed Action & 180.47 & 0.52 & 2.61 & 0.39 & 132.89 & 0.89 & 4.28 & 0.67 & 247.72 & 0.00 & 0.25 & 0.00 \\
CMA & 141.68 & 2.30 & 10.02 & 2.16 & 102.29 & 3.57 & 14.26 & 3.33 & 197.35 & 0.50 & 4.03 & 0.50 \\
TravelUAV & 138.80 & 4.18 & 20.77 & 3.84 & 102.94 & 4.63 & 22.82 & 4.24 & 189.46 & 3.53 & 17.88 & 3.28 \\
NavFoM & 125.10 & 6.30 & 18.95 & 5.68 & 102.41 & 6.77 & 20.07 & 6.04 & 170.58 & 5.36 & 15.71 & 4.97 \\
LongFly & 108.32 & 11.27 & 30.27 & 9.32 & 78.56 & 12.96 & 34.31 & 10.32 & 148.10 & 9.02 & 24.88 & 7.98 \\
AeroVLA & \textbf{67.42} & \underline{37.58} & \underline{52.92} & \underline{28.22} & \textbf{44.99} & \textbf{41.89} & \underline{58.47} & \underline{29.72} & \textbf{99.11} & \underline{31.49} & \underline{45.09} & \underline{26.11} \\
\midrule
\textbf{FreqNav (Ours)} & \underline{72.91} & \textbf{39.04} & \textbf{55.11} & \textbf{30.50} & \underline{53.90} & \underline{40.29} & \textbf{58.82} & \textbf{29.88} & \underline{99.77} & \textbf{37.28} & \textbf{49.87} & \textbf{31.38} \\
\bottomrule
\end{tabular}
}
\caption{Comparison on the Test Unseen Map Set. SR, OSR, and SPL are reported in percentage (\%). Bold and underline denote the best and second-best model results, respectively.}
\label{tab:unseen_map_full_comparison}
\end{table*}

\subsection{Experimental Setup}
\label{sec:experimental_setup}

\paragraph{Dataset.}
\label{sec:dataset}
We evaluate on TravelUAV benchmark~\cite{wang2025traveluav}, specifically the UAV-Need-Help task for object-oriented aerial navigation in AirSim~\cite{shah2017airsim}.
The benchmark contains 12,149 human-operated trajectories over 89 object categories, covering diverse outdoor scenes (\emph{e.g.,} urban, snowy, and meadow environments) and with target distances spanning 50--400\,m.
Trajectories are further grouped by length into Easy ($<250$\,m) and Hard ($\geq250$\,m).
Following the official split protocol~\cite{wang2025traveluav}, we train on 7,922 trajectories and evaluate on three held-out sets: 1,418 trajectories for Seen Scenarios, 629 for Unseen Object, and 958 for Unseen Map.

\paragraph{Evaluation Metrics.}
We report four navigation metrics.
Success Rate (SR) is the percentage of episodes that terminate within the 20\,m success radius.
Oracle Success Rate (OSR) is the percentage of trajectories that enter the success radius at any point before termination.
Success weighted by Path Length (SPL) discounts successful episodes by excess path length.
Navigation Error (NE) is the final Euclidean distance to the target in meters.
Higher SR, OSR, and SPL are better, while lower NE is better.

\paragraph{Baselines.}
\label{sec:baselines}
We compare FreqNav against recent baselines: (1) Random Action and Fixed Action are heuristic lower bounds that sample random motions or map instructions to fixed movements. (2) CMA is a recurrent cross-modal attention VLN policy adapted in prior UAV-VLN evaluations~\cite{anderson2018vision}. (3) TravelUAV-DA is the official benchmark baseline~\cite{wang2025traveluav}. (4) NavFoM and LongFly represent recent open-world and long-horizon UAV navigation approaches~\cite{zhang2025embodied,jiang2025longfly}. (5) AeroVLA is the strongest end-to-end VLA baseline on the same official splits~\cite{xu2026aerialvla}. All results are reported from original papers.

\paragraph{Implementation Details.}
\label{sec:implementation}
We use the Qwen2.5-VL-3B-Instruct~\cite{bai2025qwen25vl} as backbone with a frozen vision tower and LoRA adaptation~\cite{hu2022lora} ($r=64$, $\alpha=128$, dropout 0.05).
The vertically stacked input has width 224 and height 448, the routed prefix contains 50 visual tokens, and the grounding suffix contains 128 queries.
We train the model for 3 epochs using AdamW with a learning rate of $10^{-4}$ and a weight decay of 0.01.
The bfloat16 distributed training is performed on 4 × NVIDIA RTX A6000 GPUs with a per-device batch size of 8, resulting in a global batch size of 32.
The DiT has the hidden dimension of 512, 8 Transformer blocks, 8 attention heads, and 16 flow-inference steps.
Unless otherwise specified, the heatmap and action horizon are set to $h_g=3$ and $K=1$, respectively.

\subsection{Main Results}
\label{sec:main_results}
In this section, we report the NE, SR, OSR and SPL on the test set across seen scenarios, unseen maps and unseen objects. For each set, Hard and Easy cases are further reported based on the trajectory length.

\paragraph{Seen Scenarios.}
As shown in Table~\ref{tab:seen_full_comparison}, FreqNav achieves the best Full-set SR and SPL, improving over AeroVLA from 47.96\% to 51.55\% SR and from 38.54\% to 41.67\% SPL.
FreqNav also obtains the second-best Full-set NE and OSR, indicating that its gains come from more successful and efficient completion rather than uniformly shorter final distance.
The advantage is most pronounced on Hard trajectories, where FreqNav improves over AeroVLA by 7.40 in SR and 5.88 in SPL and also achieves the best OSR.
These results suggest that adaptive frequency routing is most beneficial when long-range search and precise terminal approach need to both be handled within the long trajectory.

\paragraph{Unseen Objects.}
As shown in Table~\ref{tab:unseen_obj_full_comparison}, FreqNav achieves state-of-the-art performance across all metrics.
Compared with AeroVLA, it improves SR from 56.60\% to 57.23\% and OSR from 64.86\% to 70.75\%, while also reducing NE from 61.45\,m to 58.39\,m.
On Easy trajectories, FreqNav matches AeroVLA's best SR and improves NE, OSR, and SPL, indicating that the routed representation preserves target recognition while improving trajectory efficiency.
On Hard trajectories, FreqNav achieves the highest SR, but LongFly has the best OSR and AeroVLA has a marginally higher SPL.
This split shows that FreqNav improves full-set unseen-object performance while leaving room for stronger landing strategy on the hardest object-generalization episodes.

\paragraph{Unseen Maps.}
As shown in Table~\ref{tab:unseen_map_full_comparison}, FreqNav achieves superior performance on unseen maps, outperforming AeroVLA by 1.46, 2.19, and 2.28, respectively.
AeroVLA retains the best Full-set NE, while FreqNav obtains the second-best NE, suggesting that FreqNav improves successful route completion more than final-distance minimization on failed episodes.
The gain is concentrated on Hard trajectories: FreqNav improves over AeroVLA by 5.79 in SR and 5.27 in SPL, with the best OSR as well.
On Easy trajectories, AeroVLA has slightly higher SR and lower NE, while FreqNav obtains the best OSR and SPL.
This pattern indicates that FreqNav is particularly useful when transfer requires robust long-horizon search in unfamiliar map layouts.

\subsection{Ablation Study}
\label{sec:ablation}

In this section, we perform several ablation studies to verify the effect of routed-token capacity, phase-dependent grounding design, and DiT action horizon.
All ablations use the same TravelUAV evaluation metrics as the main results.

\begin{table}[t]
\centering
\begin{tabular}{cccccc}
\toprule
\textbf{Routed} & \textbf{All} & \multirow{2}{*}{\textbf{NE$\downarrow$}} & \multirow{2}{*}{\textbf{SR$\uparrow$}} & \multirow{2}{*}{\textbf{OSR$\uparrow$}} & \multirow{2}{*}{\textbf{SPL$\uparrow$}} \\
\textbf{Tokens} & \textbf{Tokens} & & & & \\
\midrule
- & 128 & 66.58 & 48.76 & 58.58 & 37.36 \\
10 & 50 & \textbf{64.16} & \textbf{51.55} & \textbf{60.79} & \textbf{41.67} \\
20 & 50 & 67.35 & 39.02 & 55.26 & 31.77 \\
\bottomrule
\end{tabular}
\caption{Ablation on routed-token capacity on the test seen set. ``-'' denotes that no router is used.}
\label{tab:routed_tokens}
\end{table}

\paragraph{Number of Routed Tokens.}
Table~\ref{tab:routed_tokens} studies how many visual positions should be assigned to the adaptive evidence bank. With only 50 visual tokens, the 10/50 router improves NE from 66.58 to 64.16, SR from 48.76\% to 51.55\%, and SPL from 37.36\% to 41.67\% compared with the uncompressed 128-token baseline. However, increasing the routed portion to 20/50 degrades all metrics, dropping SR to 39.02\% and SPL to 31.77\%. This indicates that visual features in 3D navigation environments contain significant spatial redundancy. Extracting too many tokens introduces environmental noise, which leads to attention distraction in the cross-attention mechanism, preventing the model from aligning with critical instruction cues. A highly sparse visual representation is more robust for goal-conditioned decision-making.

\begin{table}[t]
\centering
\setlength{\tabcolsep}{10pt}
\begin{tabular}{ccccc}
\toprule
\textbf{Model} & \textbf{NE$\downarrow$} & \textbf{SR$\uparrow$} & \textbf{OSR$\uparrow$} & \textbf{SPL$\uparrow$} \\
\midrule
- & 82.86 & 34.69 & 48.46 & 25.15 \\
$h_g=1$ & 78.90 & 32.71 & 44.05 & 26.41 \\
$h_g=2$ & 70.64 & 48.57 & 52.85 & 38.59 \\
$h_g=3$ & \textbf{64.16} & \textbf{51.55} & \textbf{60.79} & \textbf{41.67} \\
$h_g=4$ & 86.69 & 36.68 & 47.57 & 30.57 \\
\bottomrule
\end{tabular}
\caption{Ablation on heatmap grounding horizon on the test seen set. ``-'' denotes that no visual grounding is used.}
\label{tab:grounding_horizon}
\end{table}

\paragraph{Grounding Horizon.}
Table~\ref{tab:grounding_horizon} evaluates the look-ahead horizon for grounding future waypoints as sub-goals.
As shown, the grounding-query supervision is most effective at $h_g=3$, which reduces NE to 64.16 and increases SR/OSR/SPL to 51.55\%/60.79\%/41.67\%. Shorter horizons ($h_g=1, 2$) limit the receptive field, lacking sufficient multi-hop reasoning capacity for complex path planning. Conversely, extending the horizon to $h_g=4$ degrades NE to 86.69 and SR to 36.68\%. This performance drop aligns with the well-known over-smoothing issue in deep attention and graph networks; excessive message passing steps cause distinct query representations to become indistinguishable, thereby diluting the specific semantic guidance of the navigation instruction.

\begin{table}[t]
\centering
\begin{tabular}{ccccc}
\toprule
\textbf{Waypoint Horizon} & \textbf{NE$\downarrow$} & \textbf{SR$\uparrow$} & \textbf{OSR$\uparrow$} & \textbf{SPL$\uparrow$} \\
\midrule
Baseline & 83.15 & 34.04 & 49.17 & 25.98 \\
$K=1$ & \textbf{64.16} & \textbf{51.55} & \textbf{60.79} & \textbf{41.67} \\
$K=2$ & 69.02 & 35.99 & 53.63 & 28.04 \\
$K=3$ & 77.60 & 32.09 & 47.38 & 25.80 \\
\bottomrule
\end{tabular}
\caption{DiT waypoint chunk ablation on the test seen set. ``Baseline'' denotes that we directly use MLP for next waypoint prediction.}
\label{tab:dit_k}
\end{table}

\paragraph{DiT Waypoint Horizon.}
Table~\ref{tab:dit_k} verifies the effect of the DiT waypoint chunk length $K$. While multi-step waypoint chunking is widely adopted in continuous robotic control, our results show a different trend in discrete topological navigation. The $K=1$ model achieves the best performance, improving over the baseline by 18.99 in NE and 17.51\% in SR. Increasing the chunk length to $K=2$ or $K=3$ consistently reduces performance. This is because state transitions between topological nodes represent significant viewpoint changes rather than micro-movements. Predicting long-horizon chunks in this discrete space exacerbates compounding errors (covariate shift), without providing extra execution benefits under the receding-horizon control setup. Thus, a single-step autoregressive formulation ($K=1$) remains the most reliable choice.


\subsection{Real-World Deployment}
\label{sec:deployment}

We further examine the practical deployability of FreqNav under an air-ground collaborative inference paradigm, where VLM inference is hosted on a ground server, and the predicted waypoint commands are executed by the UAV.

\paragraph{Protocol.}
We deploy the system on a Feisi J310 UAV in outdoor environments.
At each replanning step, the UAV streams front-view and down-view RGB images together with current position to a ground server.
The server runs the FreqNav policy and predicts a body-frame waypoint (\emph{i.e.,} horizontal displacement, vertical displacement, yaw angle), and stop probability.
The flight controller then executes the waypoint while enforcing flight-envelope constraints.
If communication exceeds a timeout or the returned waypoint violates the safety envelope, the UAV hovers and yields to operator takeover.

\paragraph{Latency.}
We report deployment latency from two aspects.
The first is the VLM inference latency on the ground server, measured from model input preparation to policy output.
The second is the latency of one complete replanning cycle (\emph{i.e.,} Full-link), including image/telemetry reception, VLM inference, response transmission, and network communication.
As shown in Table~\ref{tab:real_deployment}, on an NVIDIA A6000 server, FreqNav achieves an average VLM inference latency of 0.17\,s/step, compared with 0.53\,s/step for AeroVLA~\cite{xu2026aerialvla} under the same server setting.
In the real air-ground deployment, FreqNav completes one full replanning cycle in 0.5\,s/step on average.
Given that the model predicts trajectory waypoints for a 5m flight within 0.5s, this replanning latency is significantly shorter than the physical flight execution time under a safe flight speed of 2 m/s, validating sufficient real‑time performance for closed‑loop UAV navigation.

\begin{table}[t]
\centering
\setlength{\tabcolsep}{8pt}
\begin{tabular}{lccc}
\toprule
\textbf{Model} & \textbf{VLM$\downarrow$} & \textbf{Comm.$\downarrow$} & \textbf{Full‑link$\downarrow$} \\
\midrule
AeroVLA & 0.53 & 0.33 & 0.86 \\
\textbf{FreqNav (Ours)} & 0.17 & 0.33 & 0.50 \\
\bottomrule
\end{tabular}
\caption{Latency (s/step) comparison under real‑world air‑ground collaborative deployment. ``Comm.'' means communication overhead.}
\label{tab:real_deployment}
\end{table}
\section{Conclusion}
\label{sec:conclusion}

In this paper, we propose FreqNav, a frequency-routing adaptive perception framework for long-horizon aerial visual language navigation. By modeling navigation as a frequency preference shift process, FreqNav dynamically reallocates a fixed budget of visual tokens, allowing the UAV to seamlessly shift its perceptual focus from low-frequency global structures during early-stage exploration to high-frequency target details during late‑stage object reaching.
Building upon this, we integrate a Phase-dependent Grounding module for explicit spatial supervision and a Diffusion Transformer action head for continuous control. Extensive experiments on the TravelUAV benchmark demonstrate that FreqNav achieves consistent state-of-the-art performance, with the most significant improvements on complex trajectories requiring both long-horizon search and precise landing.
Finally, real-world deployments confirm the practical effectiveness of our framework. As current real‑world evaluations are restricted to a limited collection of objects and maps, future work will expand to more diverse scenarios and further validate the model’s generalization performance.

\bibliography{aaai2027}
\end{document}